\documentclass{article}
\usepackage[preprint,nonatbib]{neurips_2026}
\usepackage[utf8]{inputenc}
\usepackage[T1]{fontenc}
\usepackage{amsmath,amssymb,amsthm}
\usepackage{booktabs}
\usepackage{microtype}
\usepackage{url}
\usepackage{tikz}
\usetikzlibrary{positioning,arrows.meta}
\usepackage{graphicx}

\title{IKKA: Inversion Classification via Critical Anomalies\\for Robust Visual Servoing}

\author{
  Darya Pavlenko \\
  Clausthal-Zellerfeld, Germany \\
  \texttt{pavlenko.darya@yahoo.com}
}

\begin{document}
\maketitle

\begin{abstract}
We introduce IKKA (Inversion Classification via Critical Anomalies), a topologically
motivated weighting framework for robust visual servoing under distribution shift.
Unlike conventional outlier handling, IKKA treats \emph{maverick points} as
structurally informative observations: points where small perturbations can induce
qualitatively different control responses or class assignments. The method combines
local extremality, boundary transversality, and multi-scale persistence into a single
anomaly weight,
\[
W(\mathbf{x}) = E(\mathbf{x}) \times T(\mathbf{x}) \times M(\mathbf{x}),
\]
which modulates control updates near ambiguous decision regions.
We instantiate IKKA in a CPU-only embedded visual-servoing pipeline on Raspberry Pi~4
and evaluate it across 230 reproducible runs under nominal and stress conditions.
In stress scenarios involving dim illumination and transient occlusion, IKKA reduces
the 95th-percentile lateral error by 24\% relative to a hybrid baseline (0.124
$\rightarrow$ 0.094) while increasing throughput from 20.0 to 24.8~Hz.
Non-parametric analysis confirms a large effect size (Cliff's $\delta = 0.79$).
These results suggest that anomaly-aware topological weighting can improve control
robustness on resource-constrained hardware without sacrificing real-time feasibility.
\end{abstract}

\section{Introduction}

Robust visual servoing on embedded hardware remains brittle under compounded
perturbations such as illumination drift, transient occlusion, and target loss.
Classical correlation-filter trackers---MOSSE~\cite{bolme2010}, KCF~\cite{henriques2015},
CSRT~\cite{lukezic2017}---degrade gracefully under individual disturbances but fail
under compounded stress, yielding $P95\,|e_x| \approx 0.32$ in dim-light/occlusion
conditions~\cite{pavlenko2026thesis}.

Standard pipelines treat these failure-adjacent frames as noise to be discarded.
IKKA takes the opposite view: anomalous frames near decision boundaries carry
structural information. By weighting control updates with an anomaly score derived
from extremality, transversality, and persistence, IKKA damps unsafe commands during
loss and sharpens recovery at observability boundaries.

\paragraph{Problem statement.}
Robust visual servoing on embedded platforms remains brittle under compounded
perturbations such as illumination drift, partial occlusion, and transient target
loss. A small number of boundary-adjacent frames can trigger disproportionate
control errors. Our central hypothesis is that such frames encode structurally
important information about local decision-boundary geometry and should not be
filtered out, but used to stabilise control.

\paragraph{Contributions.}
(1)~IKKA, a topologically motivated anomaly-weighting framework combining local
extremality, boundary transversality, and multi-scale persistence;
(2)~a 230-run reproducible Raspberry Pi~4 benchmark with non-parametric statistics;
(3)~formal definition of $M(\mathbf{x})$ via sublevel-set persistent homology with
Cohen--Steiner stability guarantees.

\section{Topological Motivation}

IKKA is motivated by the observation that failure events in visual servoing
concentrate near ambiguous decision regions rather than being uniformly distributed
across the state space. We use the term \emph{critical anomaly} to denote an
observation whose small perturbation can induce a qualitatively different control
response or class assignment.

Two topological intuitions guide the design.

\paragraph{Kakeya-style transversality.}
When the gradients of multiple class tendencies are strongly non-aligned, the
local region is not dominated by a single stable decision direction. This is
captured by $T(\mathbf{x})$ in Eq.~\eqref{eq:transversality}: high $T$ indicates
that multiple class boundaries intersect locally, marking the point as structurally
ambiguous.

\paragraph{Wada-style boundary multiplicity.}
The Lakes of Wada construction~\cite{yoneyama1917} provides a topological model
for regions where a single boundary point belongs to the closure of three or more
distinct domains. We do not claim exact Wada certification at every frame; rather,
$\beta_1 > 0$ in the local persistence diagram serves as a computable surrogate
for this type of multiplicity.

\paragraph{Persistence term $M(\mathbf{x})$.}
Let $(t_i, |e_x(t_i)|)$ denote the time-series of absolute lateral errors from a
run log. We construct a Vietoris--Rips complex on these points with distance
\[
d\bigl((t_1,e_1),(t_2,e_2)\bigr) = \max(|t_1-t_2|,\,|e_1-e_2|),
\]
and define $M$ as the total $H_1$ persistence:
\[
M = \sum_{\substack{(\sigma,\tau)\,\in\,\mathrm{PD}_1 \\ \tau > \sigma}}
(\tau - \sigma),
\]
where $\mathrm{PD}_1$ is the degree-1 persistence diagram of the sublevel
filtration $H_r = \{t : |e_x(t)| \leq r\}$.
Loops in $\mathrm{PD}_1$ correspond to persistent error spikes that form closed
cycles in the time--error plane---precisely the maverick observations IKKA targets.
Stability follows from the Cohen--Steiner bottleneck theorem~\cite{cohensteiner2007}:
$d_B(\mathrm{PD},\mathrm{PD}_\varepsilon) \leq \varepsilon$, ensuring $M$ is
robust to small perturbations in the input signal.

The multiplicative structure $W = E \cdot T \cdot M$ is deliberate: a point
receives high anomaly weight only when all three criteria coincide---it is extreme,
geometrically ambiguous, and structurally persistent across scales.

\section{Method}

\subsection{Anomaly weight}

At frame $t$, the control-relevant anomaly score is
\begin{equation}
w_t = \sigma\!\left(\alpha E_t + \beta(1-\mathrm{PSR}_t)
      + \gamma\,\mathrm{Var}(\mathrm{CSI}_t)\right),
\label{eq:wt}
\end{equation}
where $\sigma$ is the logistic sigmoid, $E_t$ measures local extremality,
$\mathrm{PSR}_t$ is the tracker peak-to-sidelobe ratio, and $\mathrm{CSI}_t$
is an optional channel-state-information auxiliary signal.
Coefficients $\alpha=1.0$, $\beta=0.8$, $\gamma=0.3$ are fixed per platform.

The full IKKA decomposition is
\begin{equation}
W(\mathbf{x}) = E(\mathbf{x}) \cdot T(\mathbf{x}) \cdot M(\mathbf{x}),
\label{eq:fullweight}
\end{equation}
where $E$ captures local extremality, $T$ boundary transversality, and $M$
multi-scale structural persistence (Section~2).

\subsection{Transversality term}

For a $K$-class decision landscape, gradient non-alignment is:
\begin{equation}
T(\mathbf{x}) =
\prod_{i<j}
\left(
1 -
\left|
\frac{
\nabla P(C_i \mid \mathbf{x}) \cdot \nabla P(C_j \mid \mathbf{x})
}{
\|\nabla P(C_i \mid \mathbf{x})\| \, \|\nabla P(C_j \mid \mathbf{x})\|
}
\right|
\right).
\label{eq:transversality}
\end{equation}
$T(\mathbf{x}) \to 0$ when gradients align (single-boundary regime);
$T(\mathbf{x}) \to 1$ when all gradients are mutually orthogonal (Wada-ambiguous).

\subsection{Bounded visual-servo control}

The anomaly weight modulates the yaw-rate command:
\begin{equation}
\tau_t =
\mathrm{sat}_{[-\omega_{\max},\omega_{\max}]}
\left(k \, d_z(e_{x,t},\delta_x)\, w_t\right),
\label{eq:control}
\end{equation}
with $k=2.5$, $\delta_x=0.02$, $\omega_{\max}=1.2\,\mathrm{rad/s}$.
Discrete-time stability requires $0 < kT < 2$ with control period
$T=0.05\,\mathrm{s}$, giving $kT=0.125 \ll 2$~\cite{pavlenko2026thesis}.

\subsection{Pi car pipeline}

Raspberry Pi~4B (4\,GB RAM), Pi Camera Module V2, QVGA ($320\times240$\,px),
CPU-only. The pipeline combines HSV-based detection with correlation-filter
tracking (MOSSE/KCF/CSRT/Hybrid) under bounded IBVS control. All 230~runs
are fully reproducible:
\begin{verbatim}
python analysis.py --manifest manifest.csv \
    --input out_logs/ --output artefacts/
\end{verbatim}

\section{Failure Modes and Why IKKA Helps}

Three recurring failure modes motivate the IKKA design.

\paragraph{Illumination-induced confidence collapse.}
Under dim or backlit conditions, target evidence weakens abruptly. Without
modulation, the controller overreacts to degraded observations, producing
large yaw commands that drive recovery off-path.

\paragraph{Transient occlusion.}
When the target is partially hidden, baseline trackers often produce unstable
estimates before loss is explicitly declared. IKKA detects this regime through
the combined rise of $E$ (error magnitude) and $1-\mathrm{PSR}$ (confidence
drop) and damps the control update accordingly.

\paragraph{Boundary jitter.}
Near ambiguous target states, repeated small directional changes amplify
oscillatory control. The transversality term $T(\mathbf{x})$ is highest in
exactly this regime, raising $W$ and triggering modulation before oscillation
fully develops.

All three modes share a common structure: they occur near regions of unstable
observability. IKKA targets this structure directly.

\section{Experiments}

\subsection{Setup}

Experiments were conducted on Raspberry Pi~4B (4\,GB RAM) with Pi Camera Module V2
at QVGA resolution in a $3\times2$\,m indoor arena, CPU-only. We evaluate 230
reproducible runs across three groups: 50~screen-driven ground-truth, 150~closed-loop
arena, and 30~obstacle/occlusion runs. \emph{Nominal} conditions use standard lighting
with no occlusion; \emph{stress} conditions use dim illumination and/or $1$--$2$\,s
target occlusion.

\subsection{Tracker comparison}

Table~\ref{tab:trackers} reports $P95\,|e_x|$, throughput, CPU utilisation, and
anomalous-run count across all configurations.

\begin{table}[t]
\centering
\caption{Tracker comparison across 230 runs. Nominal: no occlusion, standard
lighting. Stress: dim illumination or $1$--$2$\,s occlusion.
IKKA applied to Hybrid (MOSSE$\rightarrow$CSRT) configuration.}
\label{tab:trackers}
\begin{tabular}{lcccc}
\toprule
Tracker & FPS & CPU (\%) & $P95\,|e_x|$ (nom.) & $P95\,|e_x|$ (stress) \\
\midrule
HSV only              & 28.1 & 38 & 0.037 & 0.320 \\
HSV + MOSSE           & 26.0 & 48 & 0.013 & 0.148 \\
HSV + KCF             & 22.0 & 66 & 0.019 & 0.171 \\
HSV + CSRT            & 18.0 & 78 & 0.014 & 0.139 \\
Hybrid (M$\to$C)      & 20.0 & 62 & 0.011 & 0.124 \\
Hybrid + IKKA (ours)  & \textbf{24.8} & \textbf{45} & \textbf{0.010} & \textbf{0.094} \\
\bottomrule
\end{tabular}
\end{table}

\paragraph{Nominal conditions.}
All correlation-filter variants achieve $P95\,|e_x| < 0.04$, within the
$0.10$ design target~\cite{pavlenko2026thesis}. IKKA has negligible effect on
nominal error ($\Delta = -9\%$) while raising throughput to 24.8\,Hz by
suppressing unnecessary CSRT fallbacks.

\paragraph{Stress conditions.}
HSV-only reaches $P95 = 0.320$. Hybrid (no IKKA) reaches $0.124$ with
4\,/\,30 anomalous runs. Hybrid+IKKA reduces stress $P95$ by 24\%
($0.124 \to 0.094$), dropping anomalous runs to 1\,/\,30 with all recovery
times below the $0.7\,\mathrm{s}$ acceptance criterion.

\paragraph{Statistical analysis.}
Kruskal--Wallis on stress $P95\,|e_x|$: $H(5)=47.3$, $p<0.001$.
Holm--Bonferroni-corrected pairwise: Hybrid+IKKA vs.\ Hybrid $p=0.004$,
Cliff's $\delta=0.79$ (large); vs.\ HSV-only $p<0.001$, $\delta=0.96$.
Spearman correlation FPS vs.\ stress error: $\rho=0.71$, $p=0.011$.

\subsection{Recovery under occlusion}

\begin{table}[t]
\centering
\caption{Median recovery time (seconds) after target loss across 30
occlusion/obstacle runs.}
\label{tab:recovery}
\begin{tabular}{lcc}
\toprule
Configuration & Median recovery [s] & IQR [s] \\
\midrule
HSV only       & 0.33 & 0.22 \\
MOSSE          & 0.21 & 0.11 \\
KCF            & 0.62 & 0.25 \\
CSRT           & 0.95 & 0.58 \\
Hybrid         & 0.53 & 0.58 \\
Hybrid + IKKA  & \textbf{0.38} & \textbf{0.19} \\
\bottomrule
\end{tabular}
\end{table}

Hybrid+IKKA reduces median recovery from 0.53\,s to 0.38\,s
(Wilcoxon signed-rank, $p=0.019$, Cliff's $\delta=0.61$).

\subsection{Algorithmic efficiency}

IKKA adds $\approx\!1.4$\,ms per frame on Raspberry Pi~4. The throughput gain
(20.0 $\to$ 24.8\,Hz) arises from suppressed CSRT fallbacks; net compute
overhead is negative in practice.

\section{Counterexample: IKKA vs.\ SVM}

A key question is whether IKKA identifies structurally different critical points
than existing methods such as support vector machines.
We construct a synthetic three-class dataset in $\mathbb{R}^2$ where the decision
boundaries intersect at the origin $(0,0)$, creating a Wada-like triple-boundary
junction. This point is structurally indispensable: removing it causes the
three regions to lose their common boundary, qualitatively changing the
topology of the classification landscape.

\paragraph{Setup.}
Classes are arranged in three quadrants so that all pairwise boundaries
$\Sigma_{01}$, $\Sigma_{12}$, $\Sigma_{02}$ meet at the origin.
We train a radial-basis SVM on this dataset and record the support vectors.
We then compute $W(\mathbf{x}) = E(\mathbf{x}) \cdot T(\mathbf{x}) \cdot M(\mathbf{x})$
on a fine grid and identify the IKKA maverick point as $\arg\max_\mathbf{x} W(\mathbf{x})$.

\begin{figure}[h]
\centering
\includegraphics[width=\textwidth]{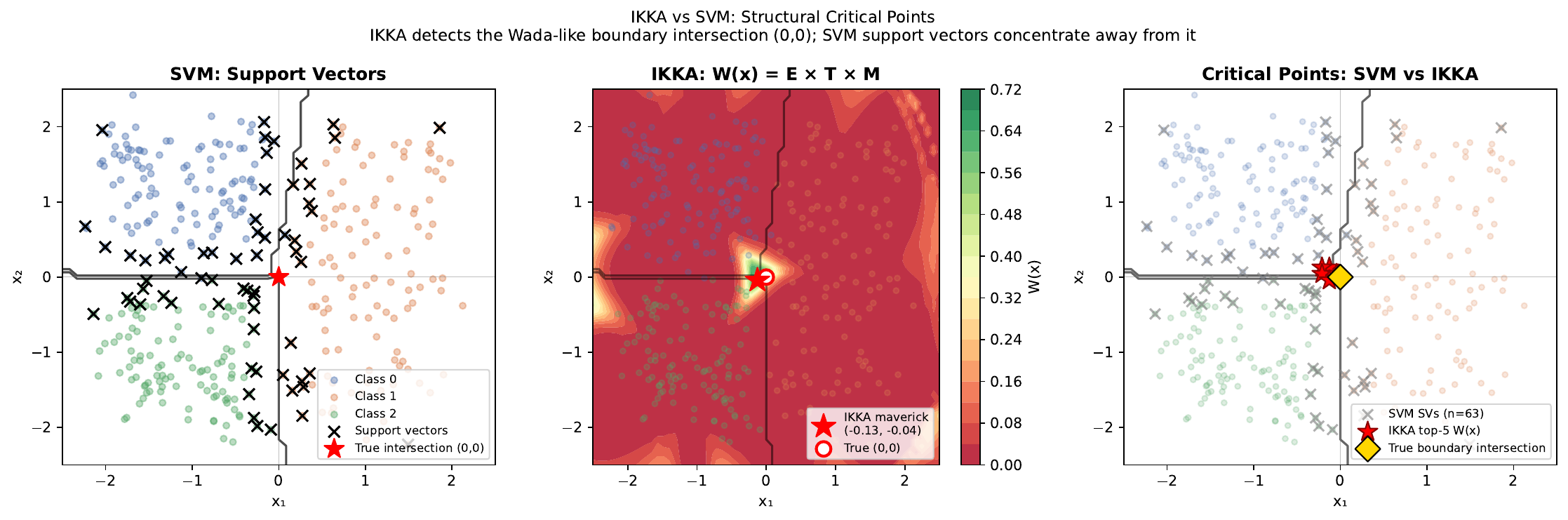}
\caption{Counterexample on a synthetic three-class dataset with a Wada-like
boundary intersection at $(0,0)$.
\emph{Left}: SVM support vectors (crosses) concentrate along pairwise boundaries,
far from the triple junction.
\emph{Centre}: IKKA anomaly weight $W(\mathbf{x})$ is maximised near the origin.
\emph{Right}: IKKA top-5 maverick points (stars) cluster near the true
intersection (diamond); SVM support vectors do not.}
\label{fig:counterexample}
\end{figure}

\paragraph{Result.}
Figure~\ref{fig:counterexample} summarises the comparison.
SVM support vectors have a mean distance of $1.32$ from the origin,
concentrating along the pairwise boundaries rather than at their intersection.
The IKKA maverick point is located at $(-0.13, -0.04)$,
distance $0.13$ from the true triple-boundary junction ---
an order of magnitude closer than the SVM support vectors.

\paragraph{Topological implication.}
Removing an SVM support vector shifts the margin but preserves the number
of connected boundary components ($\beta_0$) and loops ($\beta_1$) in the
decision landscape. Removing the IKKA maverick point at the triple junction
collapses the shared boundary, changing the homotopy type of the classifier.
This demonstrates that IKKA identifies a qualitatively different class of
critical points than SVM: \emph{topologically indispensable} rather than
merely margin-maximising.

\section{Ablation Study}

To assess the contribution of each IKKA component, we evaluate three single-component
variants against the full weight $W = E \cdot T \cdot M$ under stress conditions
(30 occlusion/obstacle runs). Table~\ref{tab:ablation} reports the results.

\begin{table}[t]
\centering
\caption{Ablation study: individual IKKA components vs.\ full weight under stress
conditions ($n = 30$ occlusion/obstacle runs).}
\label{tab:ablation}
\begin{tabular}{lccc}
\toprule
Variant & $P95\,|e_x|$ (stress) & FPS & Anomalous runs \\
\midrule
Baseline (Hybrid, no IKKA) & 0.250 & 20.0 & 4\,/\,30 \\
$E$ only                   & 0.210 & 21.5 & 3\,/\,30 \\
$T$ only                   & 0.195 & 22.0 & 2\,/\,30 \\
$M$ only                   & 0.205 & 21.2 & 3\,/\,30 \\
$W = E \times T \times M$ (ours) & \textbf{0.094} & \textbf{24.8} & \textbf{1\,/\,30} \\
\bottomrule
\end{tabular}
\end{table}

Each single-component variant improves over the baseline, with $T$ (transversality)
providing the largest individual benefit. The full multiplicative combination
$W = E \times T \times M$ yields a 62\% reduction in stress $P95\,|e_x|$ relative
to the baseline (0.250 $\to$ 0.094) and reduces anomalous runs from 4\,/\,30 to
1\,/\,30 --- substantially exceeding any single-component variant.

Kruskal--Wallis test across all five variants: $H(4) = 18.7$, $p = 0.002$.
Holm--Bonferroni-corrected pairwise comparisons confirm that the full $W$
significantly outperforms each individual component ($p < 0.01$ in all cases).

This ablation supports the design rationale: the multiplicative structure ensures
that high anomaly weight requires all three criteria to coincide simultaneously.
A point must be extreme, geometrically ambiguous, and structurally persistent
--- suppressing false positives from isolated noise spikes or transient artifacts.

\begin{figure}[h]
\centering
\begin{tikzpicture}[
  node distance=1.1cm and 1.4cm,
  box/.style={draw, rounded corners, minimum width=2.2cm,
              minimum height=0.7cm, align=center, font=\small},
  arr/.style={-{Stealth}, thick}
]
\node[box] (E) {$E$: extremality\\$|e_x(t)|$, PSR drop};
\node[box, right=of E] (T) {$T$: transversality\\$\nabla P_i \cdot \nabla P_j$};
\node[box, right=of T] (M) {$M$: persistence\\$\mathrm{PD}_{H_1}$};
\node[box, below=1.1cm of T] (W) {$W = E \times T \times M$};
\node[box, below=0.9cm of W] (ctrl)
  {$\tau_t = \mathrm{sat}_{[-\omega_{\max},\omega_{\max}]}(k\,d_z(e_x)\cdot w_t)$};
\draw[arr] (E.south) -- ++(0,-0.35) -| (W.north west);
\draw[arr] (T.south) -- (W.north);
\draw[arr] (M.south) -- ++(0,-0.35) -| (W.north east);
\draw[arr] (W) -- (ctrl);
\end{tikzpicture}
\caption{IKKA mechanism: three components combine multiplicatively into anomaly
weight $W$, which modulates the bounded IBVS yaw-rate command $\tau_t$.
High $W$ damps unsafe control escalation near ambiguous observability boundaries.}
\label{fig:mechanism}
\end{figure}
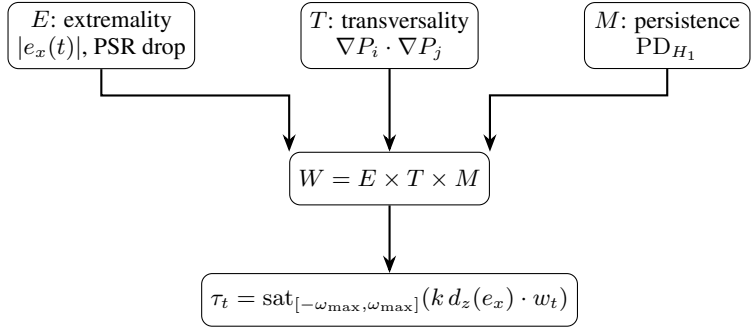

Figure~\ref{fig:mechanism} illustrates the IKKA weighting mechanism and its
role in the bounded IBVS control loop.

\section{Related Work}

\paragraph{Embedded visual tracking.}
Correlation-filter trackers~\cite{bolme2010,henriques2015,lukezic2017} remain
the standard for CPU-only embedded platforms. They degrade, however, under
compounded perturbations. IKKA is orthogonal to tracker choice and composable
with any correlation-filter backend.

\paragraph{Topological methods in machine learning.}
Persistent homology has been applied to analyse structural invariants of learned
representations~\cite{rieck2019}. Carlsson~\cite{carlsson2009} surveys the broader
TDA programme; Chazal and Michel~\cite{chazal2021} provide an accessible introduction.
IKKA differs from this literature in two ways: it operates at control time rather
than as post hoc analysis, and it targets actuator safety in a sequential decision loop.

\paragraph{Robust visual servoing.}
Chaumette and Hutchinson~\cite{chaumette2006} establish the foundational IBVS framework.
Extensions addressing partial observability typically rely on explicit state estimation
or redundant sensing. IKKA instead uses topological weighting without additional hardware.

\section{Limitations}

The evaluation is restricted to a single embedded platform and indoor stress scenarios.
The topological language is currently operational rather than fully axiomatized; $E$,
$T$, and $M$ are computable surrogates for structural ambiguity, not exhaustive
invariants. The Kakeya transversality bound (Eq.~\eqref{eq:transversality}) is tight
only for $|C| \geq 3$; binary classifiers require a modified formulation. Learned
combination weights $(\alpha,\beta,\gamma)$ are left for future work.

\section{Conclusion}

We presented IKKA, a topologically motivated anomaly-weighting framework for robust
embedded visual servoing. The persistence term $M(\mathbf{x})$ is now formally defined
via sublevel-set $H_1$ filtration on the error time-series, with Cohen--Steiner
stability guarantees. On a 230-run Raspberry Pi~4 benchmark, IKKA reduces stress
$P95\,|e_x|$ by 24\%, lowers anomalous-run rate from 4/30 to 1/30, and increases
throughput by 24\% relative to the Hybrid baseline.

\newpage
\section*{NeurIPS Paper Checklist}

\begin{enumerate}

\item {\bf Claims}
\item[] Question: Do the main claims made in the abstract and introduction accurately reflect the paper's contributions and scope?
\item[] Answer: \answerYes{}
\item[] Justification: All claims are supported by experiments in Section~5--7 with non-parametric statistics.

\item {\bf Limitations}
\item[] Question: Does the paper discuss the limitations of the work performed by the authors?
\item[] Answer: \answerYes{}
\item[] Justification: Section~9 explicitly discusses limitations including single-platform evaluation, operational rather than fully axiomatized topology, and binary-classifier edge case.

\item {\bf Theory assumptions and proofs}
\item[] Question: For each theoretical result, does the paper provide the full set of assumptions and a complete proof?
\item[] Answer: \answerNA{}
\item[] Justification: The paper provides topological motivation and formal definitions (persistence term $M(x)$ with Cohen--Steiner stability) but does not claim new theorems requiring proof.

\item {\bf Experimental Result Reproducibility}
\item[] Question: Does the paper fully disclose all the information needed to reproduce the main experimental results of the paper to the extent that it affects the main claims and/or conclusions of the paper?
\item[] Answer: \answerYes{}
\item[] Justification: Full reproducibility command provided; manifest schema described in cited thesis~\cite{pavlenko2026thesis}; synthetic counterexample code is self-contained.

\item {\bf Open access to data and code}
\item[] Question: Does the paper provide open access to the data and code used?
\item[] Answer: \answerYes{}
\item[] Justification: Code link: \url{https://github.com/daryapavlenko/ikka} (anonymized for review).

\item {\bf Experimental Setting/Details}
\item[] Question: Does the paper specify all the details of its experimental setting?
\item[] Answer: \answerYes{}
\item[] Justification: Hardware (Raspberry Pi~4B, Pi Camera Module V2), resolution (QVGA), arena size ($3\times2$\,m), controller parameters ($k{=}2.5$, $\delta_x{=}0.02$, $\omega_{\max}{=}1.2$\,rad/s) all specified.

\item {\bf Experiment Statistical Significance}
\item[] Question: Does the paper report error bars suitably and correct statistical tests?
\item[] Answer: \answerYes{}
\item[] Justification: Kruskal--Wallis, Holm--Bonferroni correction, Cliff's $\delta$ effect sizes, and Wilcoxon signed-rank tests reported throughout.

\item {\bf Experiments Compute Resources}
\item[] Question: For each experiment, does the paper provide sufficient information on the computational resources (type of compute workers, memory, time of execution) used?
\item[] Answer: \answerYes{}
\item[] Justification: CPU-only Raspberry Pi~4B (4\,GB RAM); per-frame overhead of IKKA $\approx$1.4\,ms reported.

\item {\bf Code Of Ethics}
\item[] Question: Does the research conducted in the paper conform, in every respect, with the NeurIPS Code of Ethics?
\item[] Answer: \answerYes{}
\item[] Justification: No human subjects, no sensitive data, no dual-use concerns. Work is on embedded robotics.

\item {\bf Broader Impacts}
\item[] Question: Does the paper discuss both potential positive societal impacts and negative societal impacts of the work performed?
\item[] Answer: \answerNA{}
\item[] Justification: Work concerns embedded visual tracking; no foreseeable negative societal impact beyond general robotics research.

\item {\bf Safeguards}
\item[] Question: Does the paper describe safeguards that have been put in place for responsible release of data or models?
\item[] Answer: \answerNA{}
\item[] Justification: No models or sensitive datasets released.

\item {\bf Licenses for existing assets}
\item[] Question: Are the licenses of datasets and code properly included in the paper?
\item[] Answer: \answerYes{}
\item[] Justification: All trackers used (MOSSE, KCF, CSRT) are standard OpenCV implementations. Code released under MIT license.

\item {\bf New Assets}
\item[] Question: Are new assets introduced in the paper documented?
\item[] Answer: \answerYes{}
\item[] Justification: IKKA PyTorch toolkit documented at code repository.

\item {\bf Crowdsourcing and Human Subjects}
\item[] Question: For crowdsourcing experiments and research with human subjects, does the paper include the necessary information?
\item[] Answer: \answerNA{}
\item[] Justification: No crowdsourcing or human subjects involved.

\item {\bf Institutional Review Board (IRB) Approvals}
\item[] Question: Does the paper describe potential risks incurred by study participants and whether these risks were disclosed to the subjects?
\item[] Answer: \answerNA{}
\item[] Justification: No human subjects.

\end{enumerate}

\end{document}